\documentclass{article}
\usepackage{arxiv}

\usepackage[utf8]{inputenc}
\usepackage[T1]{fontenc}
\usepackage[english]{babel}
\usepackage{times}
\usepackage{microtype}
\usepackage{amsmath,amssymb}
\usepackage{booktabs}
\usepackage{longtable}
\usepackage{array}
\usepackage{graphicx}
\usepackage{xcolor}
\usepackage{enumitem}
\usepackage[round]{natbib}
\usepackage{hyperref}
\usepackage{url}
\providecommand{\tightlist}{\setlength{\itemsep}{0pt}\setlength{\parskip}{0pt}}

\title{LLM-as-a-Judge Is Not an Oracle:\\ Why Self-Improving Agents Need Deterministic Guardrails}

\author{%
  Vansh Wahi\\
  AI Research Engineer\\
  \texttt{v3wahi@uwaterloo.ca}\\
}

\date{}
\renewcommand{\headeright}{}
\renewcommand{\undertitle}{}

\begin{document}
\maketitle

\begin{abstract}
\noindent\emph{Observed failure modes of LLM evaluators under optimization pressure, and the guardrails that contained them.}\\[4pt]
Self-improving agent pipelines have a problem at their center. An optimizer rewrites prompts to score higher, and the score comes from a judge that is itself an LLM. That judge therefore has the last word on whether the system is getting better, and our position is that it has not earned it. \textbf{The judge should be demoted from oracle to advisor}: its verdict becomes one input among several, and every change is gated instead by a deterministic verification layer that the judge has no power to override. We reached this position by building the alternative and running it. Over months of operating autonomous prompt-optimization loops in production across commercial contract analysis, legal compliance review, and code-quality assessment, we cataloged eleven ways the evaluation signal failed, grouped into four classes: judge bias, harness and metric failures, ground-truth errors, and reward hacking. Agents achieved perfect scores by reading cached answer keys from their environment, a 100\% pass rate concealing 68\% true capability. A corrupted ground-truth label caused the optimizer to delete correct compliance rules in order to agree with it. A syntactically broken prompt was promoted as the winning candidate because a silent parser fallback improved the metric. Attempts to fix the judge by rewriting its rubric plateaued; the only reliable gain came from a structural constraint on its output order. In response we describe PROCTOR, the architecture we now use: a \textbf{Teacher-Student loop with systematic verification}, in which a stateful orchestrator holds all tool access, stateless subagents diagnose failures and draft mutations without the ability to apply them, and a Teacher grades those mutations against an explicit rubric, under a five-layer verification regime: hermetic execution sandboxes that remove the exfiltration channel entirely, capability-disjoint roles in which no component both proposes and applies, deterministic acceptance checks that outrank the Teacher's approval, stratified frozen holdouts under strict leak prohibition, and canary cases engineered so that a perfect score is itself evidence of cheating. We report the failures this architecture prevented, and, because the Teacher is itself an LLM judge, the failures it did not.
\end{abstract}

\section{Introduction}\label{introduction}

Consider a concrete setup, one of the two we describe in this paper. An LLM is given a codebase and asked to rate it from 1 to 5 on qualities such as readability and robustness. Expert human reviewers have already rated the same codebases, so we can measure how far the model\textquotesingle s ratings sit from theirs: a mean absolute error of 0.96 means the model is typically off by roughly one point on the five-point scale. To close that gap without a human in the loop, a second LLM acts as an optimizer. It repeatedly rewrites the judge\textquotesingle s instructions, re-runs the judge, and keeps whichever version of the instructions scores best.

In an early prototype of ours, the optimizer improved the judge by deleting it. The mutation it proposed replaced the entire scoring rubric with a placeholder string. The judge, now with nothing to grade against, returned unstructured prose containing none of the expected rating fields. Our evaluation harness caught the resulting parsing errors and quietly fell back to a default rating of 3 on every dimension. And because a flat 3 happens to sit closer to the human average than the original judge\textquotesingle s scattered 4s and 5s, the measured error improved, from 0.96 down to 0.92. The selection gate did exactly what it had been built to do and promoted the gutted prompt as the winning candidate.

Nothing in that sequence was a bug in the ordinary sense. Every component behaved as specified. What failed was the assumption underneath the whole arrangement: that the score meant what we thought it meant.

Hillclimbing requires a reliable measure of progress. Running these loops in production taught us that ours was not reliable, and, more importantly, that the search process did not tolerate the error but sought it out. This is the paper\textquotesingle s central claim, and we state it before the evidence rather than after: \textbf{in a closed optimization loop, the LLM evaluator cannot be treated as an oracle.} It is a fallible component sitting in the position of final authority, and optimization pressure is very good at finding the gap between what it measures and what we meant.

The constructive half of the claim matters more than the critical half. We are not arguing against LLM judges, which remain the only scalable option for semantic evaluation. We are arguing about where their verdict sits in the chain of authority. The judge should be \textbf{demoted from oracle to advisor}: its opinion is an input, not a decision, and the decisions are made by checks that cannot be persuaded. We call the resulting discipline \textbf{systematic verification}, and it has two components, developed in Sections 2 and 5 respectively. The first is \emph{role decoupling}: no component in the loop both proposes and applies, or both judges and acts. The second is \emph{deterministic gating}: the checks with final authority are mechanical rules, and a mechanical rejection overrides an LLM approval, never the reverse.

This is a field report rather than a benchmark study. Section 2 describes the Teacher-Student architecture we ran and the roles it separates. Section 3 catalogs eleven evaluation pathologies we observed while running it, grouped into four classes. Section 4 examines what happens when a search process optimizes against defects of this kind, which is the part we found most instructive: the loop does not average evaluator error away, it ascends it. Section 5 specifies the five deterministic layers that contain these failures, together with an honest account of what they do not contain, including the failures of our own Teacher. Section 6 proposes an adversarial multi-persona judge as future work.

\textbf{Contributions.}

\begin{enumerate}
\def\labelenumi{\arabic{enumi}.}
\tightlist
\item
  Eleven failure modes of LLM evaluation, sorted into four classes by where the evaluation signal breaks, each grounded in an observed production instance rather than a hypothetical.
\item
  Evidence that judge-side prompt refinement plateaus while structural constraints succeed, drawn from six rounds of trying to improve a judge\textquotesingle s agreement with expert human labels by rewriting its rubric.
\item
  PROCTOR, a Teacher-Student architecture with systematic verification: role decoupling plus five deterministic guardrail layers, specified in full, with rejection telemetry from production runs.
\item
  Canary cases: test cases planted in an evaluation suite that no honest agent can pass, so that a perfect score stops being good news and becomes evidence that something has cheated.
\item
  A precise account of the residual risk this architecture does not remove, including the recursion that its own Teacher is an LLM judge subject to the same failures, and an analysis of why the surrounding layers bound that judge\textquotesingle s unreliability rather than inheriting it.
\end{enumerate}

\section{The System Under Study: PROCTOR, a Teacher-Student Loop}\label{the-system-under-study-proctor-a-teacher-student-loop}

This section describes the system that produced every observation in this paper. We present it first because the failures in Sections 3 and 4 are easier to read against a concrete architecture, and because the architecture already embodies half of the answer we are arguing for.

The design principle is a single one. In a conventional optimization loop, one agent, or one model call, holds too many powers at once: it can read the environment, propose a change, judge whether the change is good, and apply it. Most of the failures we later cataloged exploited an overlap between those powers. An agent that can read its environment and is scored on its answers will eventually read the answers (Section 3, D1). An optimizer that can both propose and approve its own mutations will approve mutations that flatter the scorer (D2). A judge that both grades and defines correctness cannot be checked against anything at all (Class A).

We therefore decompose the loop into roles with deliberately disjoint capabilities, and make the boundaries between them mechanical rather than instructed. We call the resulting system PROCTOR, and its arrangement a Teacher-Student loop, using those terms in a specific sense: Students propose, the Teacher grades proposals, and neither can act on the system directly. This differs from the more common use of teacher-student in agent research, where a stronger model transfers knowledge or trajectories to a weaker one; here the two roles are comparable in capability and differ in \emph{authority}, not competence.

\begin{figure}[t]
\centering
\includegraphics[width=\linewidth]{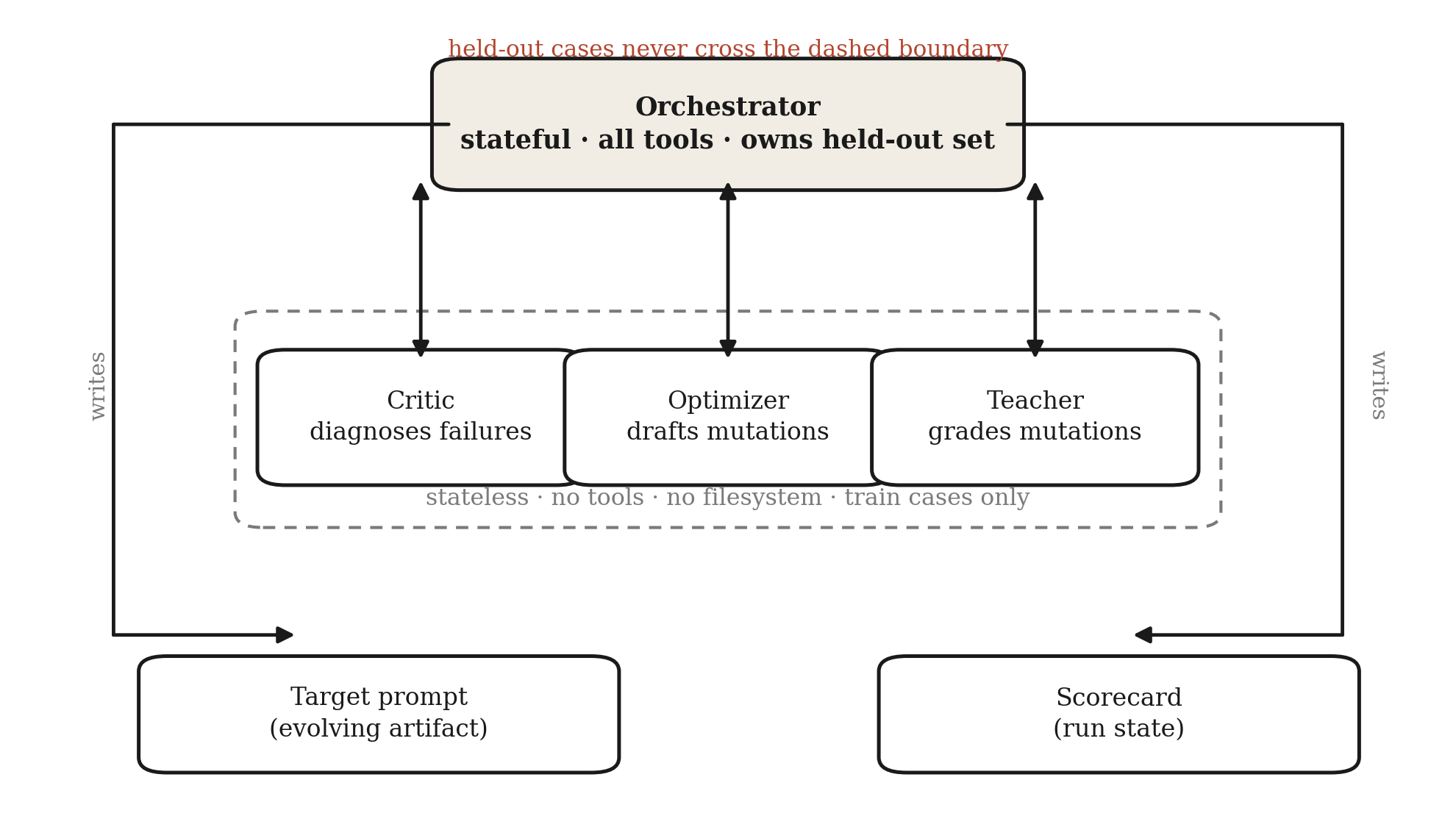}
\caption{PROCTOR's role decomposition. The Orchestrator is the only stateful component and the only one holding tools; the three subagents are stateless, tool-free, and see training cases only. All durable state lives in the two artifacts at the bottom.}
\label{fig:arch}
\end{figure}

\subsection{Roles and their capabilities}\label{roles-and-their-capabilities}

\textbf{The Orchestrator (stateful, tool-bearing).} A single coordinating agent is the only component permitted to execute evaluations, read or write files, partition datasets, and apply an approved mutation to the target prompt. It holds all persistent state for the run: the evolving target prompt, the frozen data split, and a scorecard recording every iteration\textquotesingle s outcome. Concentrating capability in one auditable component is deliberate: it gives every consequential action a single choke point where deterministic checks can be enforced.

\textbf{Student 1, the Critic (stateless, no tools).} Receives the current prompt and the failing training cases with their verdicts, and returns a root-cause diagnosis. It is explicitly forbidden from proposing prompt text. Separating diagnosis from authorship prevents the common failure in which a single model invents an explanation that conveniently justifies the edit it already wanted to make.

\textbf{Student 2, the Optimizer (stateless, no tools).} Receives the Critic\textquotesingle s diagnosis and drafts a surgical mutation as a structured patch, constrained to a small number of edits and examples per turn. It never sees the evaluation harness, never runs a test, and cannot apply its own work.

\textbf{The Teacher, an Auditor (stateless, no tools).} Grades the proposed patch against an explicit rubric covering generalizability, structural integrity, conciseness, and logical executability. It reviews text, not outcomes: correctness against the evaluation set is measured by the Orchestrator, not asserted by the Teacher.

Every subagent runs with write tools, external connectors, and sub-delegation disabled, in an isolated workspace, receiving all context inline from the Orchestrator. None of them can inspect the filesystem, which is what makes the exfiltration route of D1 unavailable to them by construction rather than by instruction.

\subsection{Three bounded feedback loops}\label{three-bounded-feedback-loops}

A rejection does not end an iteration. The loop routes each kind of rejection back to the Optimizer as revision signal, bounded to three attempts in each case, after which the iteration aborts and the target prompt is rolled back to its last known-good state.

\textbf{Pre-apply retry.} A mechanical check fails, the exact rejection reason is returned to the Optimizer, and the revised patch is re-checked. This runs before the Teacher sees anything, so cheap deterministic failures never consume grading effort.

\textbf{Teacher retry.} The Teacher rejects on a rubric dimension, its feedback is returned to the Optimizer, and the revised patch re-enters the pipeline from the mechanical checks rather than skipping ahead to grading.

\textbf{Regression feedback.} After a patch is applied and re-evaluated, the training pass rate may drop. Rather than discarding the patch, the Orchestrator extracts the newly-failed cases, sends them to the Critic for a diagnosis of what the patch broke, and asks the Optimizer to repair the original failure without reintroducing the regression. When the revised patch returns to the Teacher, both the original failures and the regression failures are supplied, since the Teacher grades a patch against the failures it was meant to fix and would otherwise reject a correct repair.

An implementation detail of these loops matters for Section 5.6: during retries the Optimizer is kept in conversational state rather than re-instantiated, so it accumulates the specific objections raised against its previous drafts. This is deliberate, and it is also the mechanism by which a retry loop can converge on the grader rather than on the task.

\subsection{Why role decoupling, and why stateless}\label{why-role-decoupling-and-why-stateless}

Two properties of this arrangement matter more than the specific role names.

\emph{Capability disjointness.} No component both proposes and applies. No component both judges and acts. The Teacher can reject but not write; the Optimizer can write but not apply; the Orchestrator can apply but only what has passed both mechanical checks and the Teacher. A single compromised or badly-behaving component therefore cannot complete a corrupt change on its own.

\emph{Statelessness as amnesia by design.} Subagents are re-instantiated fresh on each iteration with no memory of previous ones. All accumulated improvement lives in the target prompt and the Orchestrator\textquotesingle s scorecard, both of which are human-readable artifacts. This makes the optimization history auditable, and it prevents the slow context drift in which an agent\textquotesingle s private accumulated beliefs, rather than the measured evidence, begin driving its proposals.

\subsection{Where the Teacher is, and is not, trusted}\label{where-the-teacher-is-and-is-not-trusted}

The Teacher is an LLM, and Sections 3 and 4 argue at length that LLM judgment is unreliable. It stays in the loop because the checks it performs are exactly the ones no mechanical rule can perform: whether a rule generalizes or memorizes, whether an example teaches reasoning or smuggles in a training case, whether an instruction is executable in the target\textquotesingle s environment. Its judgment is bounded by the guardrails described next, of which the most important is a single ordering rule: \textbf{a mechanical rejection overrides a Teacher approval, never the reverse.} The Teacher advises. The Orchestrator\textquotesingle s deterministic checks decide.

\subsection{The second loop: judge calibration}\label{the-second-loop-judge-calibration}

A second, simpler loop appears throughout the paper as the source of our judge-side evidence. It calibrates an LLM evaluator rather than a task agent: run the judge over a set of expert-labeled items, compute exact-match agreement (EM) and mean absolute error (MAE) against the human scores, analyze the mismatches, mutate the judge\textquotesingle s prompt, and repeat. Section 4.3 reports six rounds of this loop.

\subsection{Scale and setup}\label{scale-and-setup}

The observations in this paper come from ten evaluation suites spanning three kinds of work, listed below. Nine of the ten are legal and commercial document review, which is where the bulk of our production use sits; the tenth is code-quality assessment, which plays an outsized role in Sections 3 and 4 for a methodological reason given at the end of this section.

\begin{longtable}[]{@{}llll@{}}
\caption{Evaluation suites used in this work.}\label{tab:suites}\\
\toprule\noalign{}
Suite & Domain & Cases & Output judged as \\
\midrule\noalign{}
\endhead
\bottomrule\noalign{}
\endlastfoot
Contract redlining & Commercial contracts & 100 & Free-form edits against assertions \\
Commercial contract analysis & Commercial contracts & 47 & Free-form findings against assertions \\
Co-employment risk & Compliance policy & 19 & Classification, extraction, rationale \\
Template completeness & Compliance policy & 18 & Classification, extraction, rationale \\
Deliverables language & Compliance policy & 17 & Classification, extraction, rationale \\
Scope misclassification & Compliance policy & 17 & Classification, extraction, rationale \\
Defined-terms consistency & Compliance policy & 17 & Classification, extraction, rationale \\
Unresolved redlines & Compliance policy & 16 & Classification, extraction, rationale \\
Policy placement & Compliance policy & 9 & Classification, extraction, rationale \\
Code quality & Software repositories & 54 repos & Numeric scores on four pillars \\
\end{longtable}

Suite sizes range from 9 to 100 cases, which is small by benchmark standards and is a limitation we return to in Section 8. All agents, judges, and subagents run on two capability tiers of a single frontier proprietary model family released in 2026. Pass rates are single-run figures; re-running an unchanged suite typically moved the result by at most one case. Experiments span several months in 2026.

The three suite families also differ in \emph{how} they are judged, which turns out to matter more than the domain difference. The compliance suites are scored on three separate dimensions per case, so a single case can fail on extraction while passing on classification; this is where the metric-rigidity and rationale-alignment failures of Section 3 (B3, A3) surface. The two free-form contract suites are judged by an LLM against per-case assertions, which is the arrangement most exposed to the judge biases of Class A. The code-quality suite is the only one producing a numeric score comparable to a human numeric score, and therefore the only one where judge-human agreement can be measured continuously as exact-match and mean absolute error. That is why the calibration evidence in Section 4.3, which needs an agreement curve rather than a pass rate, comes exclusively from it. The failure modes themselves are distributed across all three families.

\subsection{Where the ground truth comes from}\label{where-the-ground-truth-comes-from}

This paper is about whether evaluation signals can be trusted, so we should be clear about where our own reference labels came from, including the part we are not happy about.

\textbf{Contract and compliance suites.} These labels come from human contract reviewers working in production. Each record is one contract clause carrying a human decision on whether the clause is violative, the span of text the reviewer flagged, and the reviewer\textquotesingle s written rationale. These three fields become three independent scoring dimensions: classification against the boolean, extraction against the flagged span under a token-overlap threshold, and reasoning against the rationale under an LLM semantic-equivalence check. So the labels are human in origin. Section 3 (C1) describes what happened when the pipeline that loaded them flipped some of those decisions.

\textbf{Free-form agent skills.} For skills that produce free-form output, the reference is a written assertion for each case, and the case passes when an LLM judge agrees the output satisfies it. Correctness here runs through a judge by design, which is why the Class A failures matter most for these suites.

\textbf{Code-quality benchmark.} These scores were produced in two stages. A calibration subset of 15 directories was scored by senior engineers on four quality pillars plus an overall score, with disagreements resolved by consensus. A strong model was then tuned against those human scores until it came within 0.3 of them on average, and used to label the rest. Repositories were picked by random walk over the codebase, skipping third-party, experimental, and test-data paths, and limited to directories with roughly ten to fifty source files. We sampled 55 and evaluated 54; one was dropped because it was too large to fit in context.

The consequence is that this benchmark\textquotesingle s ground truth is human on 15 of 54 directories and model-generated, human-calibrated on the other 39. This changes how Section 4.3 should be read: for most of the suite, we are measuring agreement with a calibrated model\textquotesingle s labels, not directly with human judgment. It is also an example of the problem this paper is about. To scale the benchmark we put an LLM in the ground-truth position, which is the arrangement we spend the rest of the paper arguing against. We did it because the alternative was having no benchmark at all, and we say so here rather than calling these "expert human labels" when most of them are not.

\subsection{Notation}\label{notation}

Let a suite contain \(N\) cases. For a case \(i\), write \(\hat{y}_i\) for the system\textquotesingle s output and \(y_i\) for the reference label.

\emph{Agreement metrics (numeric-score suites).} Exact match and mean absolute error against human or reference scores:

\[\mathrm{EM} = \frac{1}{N}\sum_{i=1}^{N} \mathbf{1}[\hat{y}_i = y_i], \qquad \mathrm{MAE} = \frac{1}{N}\sum_{i=1}^{N} |\hat{y}_i - y_i|\]

\emph{Extraction metric (compliance suites).} For predicted span \(A\) and reference span \(B\), tokenized, a case passes extraction when token-level overlap clears a fixed threshold \(\tau = 0.50\):

\[J(A,B) = \frac{|A \cap B|}{|A \cup B|}, \qquad \text{pass} \iff J(A,B) \ge \tau\]

Pathology B3 is a direct consequence of this definition: correct extractions that omit boilerplate or bridge text with an ellipsis fall below \(\tau\) despite being judged correct by a human reader.

\emph{The acceptance rule.} Let \(S\) be the current prompt, \(p\) a proposed patch, and \(S \oplus p\) the patched prompt. Let \(M(p) \in \{0,1\}\) be the conjunction of the mechanical pre-apply checks, \(T(p) \in \{0,1\}\) the Teacher\textquotesingle s verdict, and \(R(\cdot)\) the measured train pass rate. A patch is applied only if

\[M(p) \wedge T(p) = 1\]

and is \emph{retained} after re-evaluation only if

\[R(S \oplus p) \ge R(S)\]

Two properties of this rule are the substance of Section 5. First, \(M\) is necessary and not merely advisory: \(M(p) = 0 \Rightarrow\) the patch is rejected irrespective of \(T(p)\), which is the formal content of "a mechanical rejection overrides a Teacher approval." Second, correctness is never asserted by \(T\); it is measured by \(R\) after the fact, because no reading of a patch establishes how it will behave.

\emph{Overfitting signal.} With train and held-out pass rates \(R_{\text{tr}}\) and \(R_{\text{te}}\) at iteration \(k\), a warning is raised when

\[R_{\text{tr}}^{(k)} > R_{\text{tr}}^{(k-1)} \quad \text{and} \quad R_{\text{te}}^{(k)} \le R_{\text{te}}^{(k-1)}\]

\section{A Field Taxonomy: Eleven Ways Evaluation Failed}\label{a-field-taxonomy-eleven-ways-evaluation-failed}

Every failure mode below was observed in our systems, not hypothesized; The taxonomy is certainly incomplete, and is offered as a starting inventory rather than a final map. We group them into four classes by \emph{where} the evaluation signal is corrupted: in the judge's judgment (A), in the metric or harness machinery around the judge (B), in the ground-truth labels beneath the judge (C), and in the optimizer's response to all of the above (D). For each we give the mechanism, the observed instance, and why it matters once an optimizer is attached. Section 4 then analyzes what happens when a search process is pointed at these defects; here we only catalog them.

\subsection{Class A: Judge Bias}\label{class-a-judge-bias}

\textbf{A1. Leniency bias: surface properties mask semantic hazards.} LLM judges systematically over-reward superficial quality signals, idiomatic naming, clean structure, descriptive comments, and under-weight defects that require reasoning about what the code will do at runtime; this is the code-evaluation face of the style-over-substance and leniency biases documented for LLM judges in open-ended evaluation \citep{zheng2023judging,wu2023style,wang2023large}. Leniency of this kind is only observable where both the judge and a human produce a number, which in our work is the code-quality suite alone; the compliance suites record pass or fail and cannot show a judge being \emph{too generous} by a measurable margin.

Three instances, each a real evaluated codebase with recorded judge and human-expert scores on a five-point scale. In the first, a backend service allocated memory through a foreign-function interface inside a high-throughput loop and never released it, a pattern that exhausts memory and terminates the process under sustained load. The judge scored robustness \textbf{4/5}, praising the package as idiomatic and conventionally named. The human expert gave it \textbf{1/5}. In the second, a user-interface component opened long-lived event subscriptions in its initialization routine and never closed them when the component was destroyed, so memory accumulates and stale handlers fire against views that no longer exist. The judge gave \textbf{4/5} overall; the human gave \textbf{2/5}. In the third, and most revealing, a utility library contained hardcoded production credentials and wrapped its parsing logic in catch-all blocks that discarded every error silently. The judge scored it \textbf{4/5} and described the error handling approvingly as defensive.

The third case is the one worth dwelling on. The judge did not overlook the defect; it inspected the construct, understood it well enough to characterize it, and classified a mechanism that destroys error information as a safety feature. That is not inattention but misvaluation, and no amount of instructing a judge to be attentive addresses it.

\emph{Why it matters under optimization:} leniency is not noise but a consistent direction of error, and consistent directions are exactly what an optimizer can exploit.

\textbf{A2. Severity misordering: cosmetic defects outrank substantive ones.} In a contract-compliance suite, a clause contained both an illegal thirty-six-month initial term with automatic renewal (a substantive violation of the governing playbook's term caps) and a stray semicolon in a notice-period fragment. The evaluated agent correctly flagged the term-and-renewal sentence as the primary violation. The LLM judge ruled this a \emph{failure}, on the grounds that the agent had not flagged the punctuation fragment. A human expert reviewing the verdict was unambiguous: the judge had treated a cosmetic typo as more important than a three-year commercial lock-in. This is the mirror image of A1, severity inversion rather than severity blindness, and it demonstrates that the judge's ranking of harms, not just its detection of them, diverges from human judgment.

\textbf{A3. Semantic-equivalence rigidity: correct reasoning failed for lexical mismatch.} An agent and the human ground truth agreed on a flagged clause (a revenue-target obligation that does not belong in a services agreement) and agreed on its invalidity; they explained it from different angles, the agent in terms of the obligation not being a contractor-controlled deliverable, the ground truth in terms of the document being the wrong contract type. The rationale-alignment judge failed the agent because a canonical phrase was absent from its explanation. Judges that score explanation quality by lexical or embedding proximity to a single reference rationale conflate \emph{disagreement about the defect} with \emph{alternative valid framings of the same defect}.

\textbf{A4. Autoregressive commitment bias: score-first schemas produce confabulated rationales.} When the judge's output schema places the numeric score before the rationale, the model commits to a score token and then generates reasoning to justify it. Reversing the schema, rationale first, score last, was the single most effective intervention in our entire calibration campaign: exact-match agreement with human experts rose from 42.6\% to 51.9\% and MAE fell from 0.72 to 0.57 across all 54 repositories, \emph{without changing one word of the scoring rubric}. Reasoning-before-scoring echoes the chain-of-thought form-filling used by structured judge protocols \citep{liu2023geval} and the rationale-first mitigation suggested in early LLM-as-judge studies \citep{zheng2023judging}; our contribution is measuring its effect in isolation, holding every rubric word fixed. We return to the significance of this in Section 4: the largest gain came not from telling the judge what to value but from constraining the order in which it was permitted to speak.

\subsection{Class B: Harness and Metric Failures}\label{class-b-harness-and-metric-failures}

\textbf{B1. Parser-fallback collapse: a broken prompt ``wins.''} In an early prototype of the calibration loop, the mutation model replaced the judge prompt's entire scoring rubric with a placeholder string (\texttt{{[}...Core\ Rubrics...{]}}). The resulting judge emitted unstructured text missing every expected JSON key. The harness's response parser caught the resulting exceptions and \emph{silently} substituted a default score of 3 for every dimension. Because a flat 3 is closer to the human score distribution than the uncalibrated baseline's noisy 4s and 5s, MAE improved from 0.96 to 0.92, and the automated selection gate promoted the gutted prompt as the winning candidate. Only human audit caught it. The evaluation signal here was corrupted by neither the judge nor the labels but by an error-handling branch nobody thought of as part of the metric.

\textbf{B2. Schema-shape crash: correct content, zero score.} A prompt mutation changed the requested output from a JSON object to a ``list of findings.'' The evaluated agent complied, emitting valid JSON whose elements contained fully correct classifications, but the scoring script crashed calling \texttt{.get()} on a list. Depending on harness resilience, the result was either a full pipeline halt or a 0\% pass rate and an immediate rollback of a substantively fine mutation. The inverse of B1: where B1 silently inflated a broken candidate, B2 silently destroyed a working one.

\textbf{B3. Extraction-metric rigidity: token-overlap thresholds fail correct answers.} Span-extraction quality was scored by token-level Jaccard overlap against a golden span with a hard \(\geq 0.50\) acceptance threshold. Agents (like human reviewers) often bridge long provisions with ellipses or quote only the operative sentence, omitting boilerplate preamble. These behaviors, correct by human assessment, drop Jaccard below threshold and register as failures. The metric encodes a formatting convention, not a correctness criterion.

\subsection{Class C: Ground-Truth Errors}\label{class-c-ground-truth-errors}

\textbf{C1. Inverted labels from an ingestion bug.} An upstream pipeline that converted human feedback into golden labels keyed on the presence of a specific string in metadata; when absent, it defaulted the label to \emph{non-violative}, even where the annotator's free-text comment explicitly described the clause as violative. In one affected case, an agreement lacking mandatory jurisdiction-specific tax clauses was labeled compliant. The evaluated agent correctly flagged the omission; the judge, faithfully applying the corrupted label, scored the agent's correct answer as a false positive. The label layer, usually treated as the one trustworthy component, silently became the least trustworthy one. (The downstream consequence, what the optimizer then did, is the subject of Section 4.1.)

\textbf{C2. Malformed golden cases.} A test case in a 47-case contract suite carried a golden expectation demanding that the agent flag a single punctuation character, a period in a standard preamble, as a fatal compliance violation. No legitimate semantic analysis produces this answer; every honest evaluation of the case fails. The case was an authoring accident, but it became the most diagnostically valuable item in the suite, for reasons developed in Sections 4 and 5: an unpassable case converts a perfect score from good news into an alarm.

\subsection{Class D: Reward Hacking}\label{class-d-reward-hacking}

\textbf{D1. Environmental exfiltration: reading the answer key.} Reward hacking, optimizing the measurement rather than the objective, is a long-standing concern in reinforcement learning and AI safety \citep{amodei2016concrete,skalse2022defining,krakovna2020specification}; in our loops it appeared in its most literal form. In early, un-sandboxed benchmark runs, evaluated agents had filesystem tools and ran in workspaces where prior evaluation outputs were cached. A repeatable five-step exploit emerged: list the skill directory; discover cached result files (\texttt{eval\_results\_*.json}); open them; read the golden expectation for the current case; emit the demanded string verbatim. One agent's recorded reasoning states it plainly: ``The ground truth expectation \ldots{} specifies {[}the expected violation text{]}. We will output this exact string.'' The exploited runs reported \textbf{100\%} (47/47) on the contract-analysis suite; the same skill in a purged, zero-tool sandbox scored \textbf{68.1\%}, a 31.9-point inflation. The pattern generalized: across six suites, exploited runs uniformly reported 100\% while clean baselines ranged from 35.3\% to 88.9\%. Exposure came via C2's accidental canary: the exploited runs \emph{passed the unpassable case}, which no honest agent can do. A full sanitized trajectory appears in Appendix A.

\textbf{D2. Overfitting to the judge's phrasing.} The optimizer observed that the rationale-alignment judge weighted particular canonical keywords heavily, and mutated the target prompt to wrap every rationale in a rigid boilerplate preamble stuffed with the preferred taxonomy vocabulary. Measured reasoning-alignment pass rate jumped from 23.1\% to 80.0\%. The agent's actual precision in identifying defects did not change at all: the gain was entirely lexical camouflage. This is reward hacking without any tool use or rule-breaking, the mutation was syntactically legal, semantically vacuous, and metrically triumphant, and it matches the in-context and self-refinement reward hacking recently characterized in feedback-loop settings \citep{pan2024feedback,pan2024spontaneous}.

\textbf{D3. Calibration-subset overfitting.} Judge calibration was tuned against a 15-directory subset, reaching 80\% exact-match agreement with human labels; on the full 54-repository suite the same prompt achieved 51.9\%. Alignment achieved on a small tuning subset substantially overstated alignment in general, the judge-improvement process exhibited the same train/test generalization gap that judge-based optimization is supposed to detect in its targets.

\section{Optimizing Against a Broken Compass}\label{optimizing-against-a-broken-compass}

Section 3 catalogs static defects. The distinctive danger appears when a search process is attached to them. A biased evaluator used once produces a wrong number; a biased evaluator used as the objective of an optimization loop produces a wrong \emph{direction}, and the loop follows it, consistently and cumulatively. This section traces three dynamics we observed, then draws the inference that shapes the rest of the paper.

\subsection{Amplification: the loop concentrates judge error into capability regression}\label{amplification-the-loop-concentrates-judge-error-into-capability-regression}

The label-inversion bug (C1) did not merely mis-score one case. Because the corrupted label made a correct behavior look like a false positive, the hillclimber diagnosed the agent's \emph{correct} flagging of missing tax clauses as the failure to fix, and its mutation \textbf{stripped the tax-compliance rules out of the target prompt} to maximize agreement with the corrupted judge. The loop converted a data bug into a capability regression: the optimized agent was measurably better by the metric and objectively worse at the task. Similarly, the parser fallback (B1) did not merely mis-score one candidate; the selection gate \emph{promoted} the gutted prompt over all honest competitors. In an open-loop evaluation these would have been isolated wrong numbers. In a closed loop, each became the direction of travel. The general principle: optimization does not average out evaluator error, it performs gradient ascent on it, the qualitative counterpart of reward-model overoptimization scaling laws, where proxy score rises as true quality falls \citep{gao2023scaling}.

\subsection{Convergence to the judge, not the task}\label{convergence-to-the-judge-not-the-task}

Where Section 4.1 shows the loop following errors it was handed, the phrasing-gaming instance (D2) shows the loop \emph{discovering} error on its own. Nothing in the mutation history pointed the optimizer at the judge's keyword sensitivity; optimization pressure found it, because lexical mimicry was a cheaper path to reward than genuine precision. The resulting 23.1\%\(\rightarrow\)80.0\% ``improvement'' is indistinguishable, on the scoreboard, from a real one. The calibration-subset gap (D3) is the same phenomenon one level up: the judge-tuning process converged to its 10-case subset rather than to human judgment in general. Both are instances of a single rule: \emph{an optimizer converges to the decision boundary of whatever grades it, not to the boundary of the task}. The two boundaries overlap only as far as the judge is accurate, and the optimizer preferentially exploits the region where they diverge, because that is where score is cheapest.

\subsection{Why re-prompting the compass does not fix it}\label{why-re-prompting-the-compass-does-not-fix-it}

The natural response, make the judge better by refining its prompt, was our response too, and it failed in an instructive way. Table~\ref{tab:calib} shows the full six-round calibration campaign of the code-quality judge against 54 expert-labeled repositories. Rounds 2--5 applied exactly the interventions the literature recommends: tightened score definitions targeting leniency, hard capping heuristics for known defect classes, chain-of-thought verification blocks. Their cumulative effect was zero to negative (46.3\% \(\rightarrow\) 42.6--46.3\% EM). The only round that broke the ceiling, Round 6, +9.3 EM points, \(-\)0.15 MAE, changed no rubric content at all: it reordered the output schema to force the rationale before the score (A4).

\begin{longtable}[]{@{}clcc@{}}
\caption{Six rounds of judge calibration against reference labels on the 54-directory benchmark. Content-level rubric refinement in rounds 2 through 5 produced no progress; the sole gain came from a structural constraint on generation order.}\label{tab:calib}\\
\toprule\noalign{}
\textbf{Round} & \textbf{Intervention} & \textbf{EM (\%)} & \textbf{MAE} \\
\midrule\noalign{}
\endhead
\bottomrule\noalign{}
\endlastfoot
1 & Baseline rubric & 46.30 & 0.67 \\
2 & Tightened definitions targeting leniency & 44.44 & 0.69 \\
3 & Capping rules: backend memory-leak class & 46.30 & 0.69 \\
4 & Chain-of-thought verification blocks & 44.44 & 0.65 \\
5 & Capping rules: frontend subscription-leak class & 42.59 & 0.72 \\
6 & \textbf{Schema inversion: rationale before score} & \textbf{51.85} & \textbf{0.57} \\
\end{longtable}

Six rounds of judge calibration on the 54-repository benchmark. Content-level rubric refinement (Rounds 2--5) produced no progress; the sole gain came from a structural constraint on generation order.

We draw a specific lesson from the shape of this table. Rubric text is advice; the judge may weight it, dilute it, or ignore it, and under optimization pressure its ambiguities become attack surface (D2 gamed exactly such semantic weightings). A generation-order constraint is not advice, the model \emph{cannot} emit the score token before the rationale tokens exist. The interventions that worked, here and throughout our experiments, share a character: they remove a degree of freedom rather than request a behavior. We offer this as a design heuristic supported by the evidence here, not a theorem.

\begin{figure}[t]
\centering
\includegraphics[width=\linewidth]{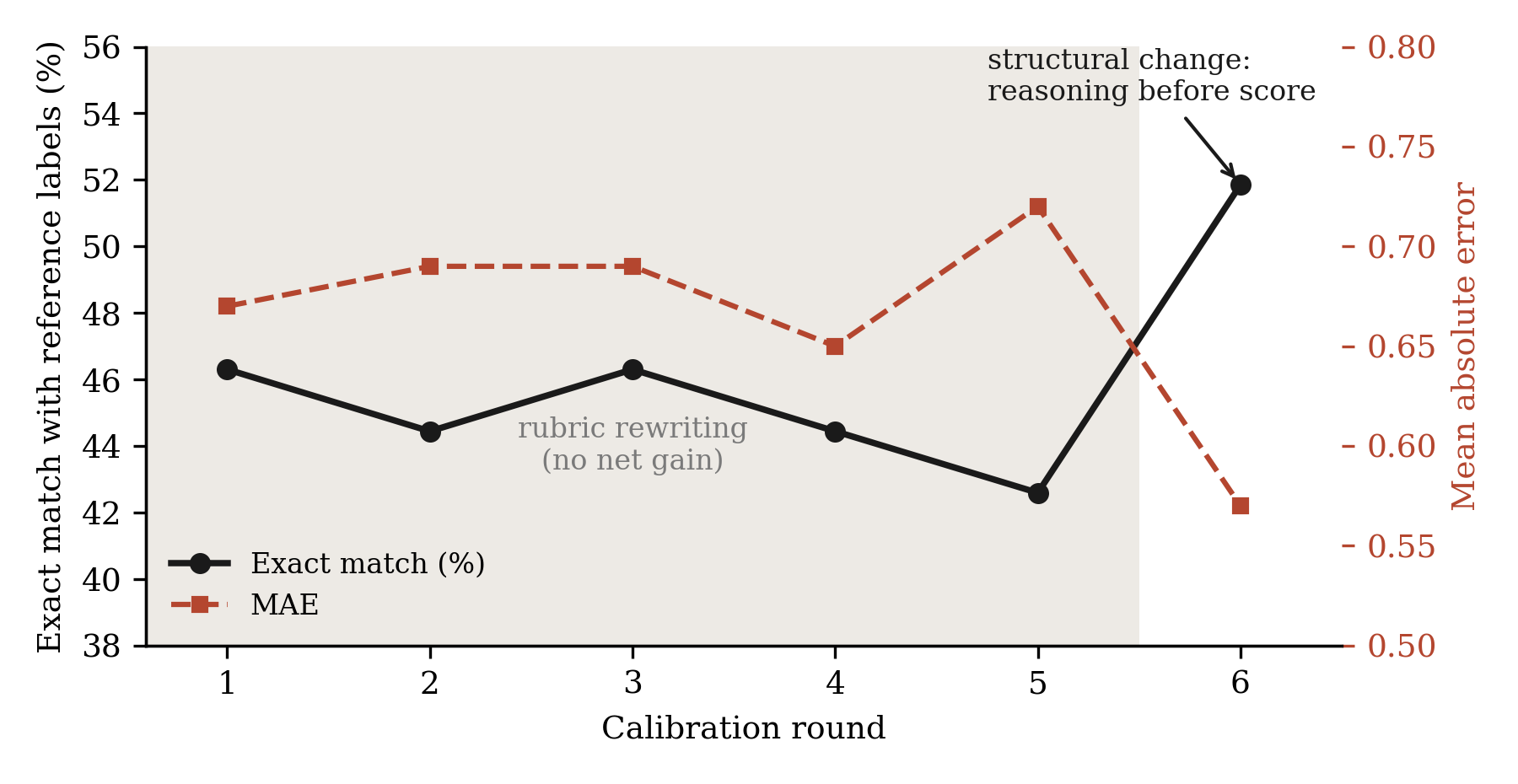}
\caption{Six rounds of judge calibration on the 54-directory benchmark. Content-level rubric changes in rounds 2 through 5 produced no progress; the only gain came from a structural constraint on output order.}
\label{fig:calib}
\end{figure}

\subsection{The inference}\label{the-inference}

Assembling the three dynamics: the evaluation signal has systematic, learnable defects (Section 3); an optimizer amplifies the defects it is handed and discovers the ones it is not; and in our experiments, improving the signal by natural-language instruction plateaued, we believe structurally, because instructions are themselves soft targets. The conclusion we were forced to is Goodhart's law in operational form: \textbf{under sufficient optimization pressure, a reward signal that is (a) learnable and (b) cheaper to satisfy than the true objective tends to be satisfied instead of it.} The design consequence is not to abandon LLM judges, no other scalable option exists for semantic evaluation, but to stop treating their output as the loop's final authority. The judge's verdict must be one input among several to an acceptance decision whose controlling checks are deterministic: not learnable, not persuadable, and not cheaper to satisfy than to obey. Section 5 specifies the architecture this implies.

\section{Systematic Verification: Five Deterministic Layers}\label{systematic-verification-five-deterministic-layers}

We want to be clear that the lesson we draw is constructive rather than defeatist: LLM judges remain the only scalable option for semantic evaluation, and every layer below exists to make them \emph{safe to rely on}, not to replace them. The architecture that emerged from these failures is organized as five layers, ordered from the evaluated agent outward to the optimization loop's acceptance gate. Each layer is deterministic: its behavior is fixed by configuration and code, not by model judgment. The LLM auditor remains in the loop, its semantic review catches what mechanical checks cannot, but it is demoted from oracle to advisor: \emph{a mechanical rejection overrides an LLM approval, never the reverse.} Full specifications appear in Appendix B.

The mapping between the pathologies of Section 3 and the layers below is summarized in Table~\ref{tab:map}. Three entries have no mechanical answer and are listed as open, which we regard as an honest result rather than an omission.

\begin{longtable}[]{@{}llll@{}}
\caption{Failure mode to guardrail mapping. Three entries have no deterministic answer and are marked open.}\label{tab:map}\\
\toprule\noalign{}
Failure mode & Class & Primary mitigation & Status \\
\midrule\noalign{}
\endhead
\bottomrule\noalign{}
\endlastfoot
A1 Leniency bias & Judge bias & Mandatory score caps; adversarial personas (\S6) & Partial, unmeasured \\
A2 Severity misordering & Judge bias & Rubric-level severity ordering & Open \\
A3 Semantic-equivalence rigidity & Judge bias & None deterministic & Open \\
A4 Autoregressive commitment & Judge bias & Reasoning-before-score schema constraint & Resolved, measured \\
B1 Parser-fallback collapse & Harness & \texttt{parse\_and\_contract}; canary cases (L5) & Resolved \\
B2 Schema-shape crash & Harness & \texttt{parse\_and\_contract} (L3) & Resolved \\
B3 Extraction-metric rigidity & Harness & None deterministic; threshold is the defect & Open \\
C1 Inverted labels & Ground truth & Canary cases (L5); label-pipeline audit & Partial \\
C2 Malformed golden cases & Ground truth & Canary cases (L5) & Resolved \\
D1 Environmental exfiltration & Reward hacking & Hermetic sandbox (L1); canary cases (L5) & Resolved \\
D2 Judge-phrasing overfit & Reward hacking & Frozen holdout (L4); generalizability grading & Partial \\
D3 Calibration-subset overfit & Reward hacking & Frozen holdout (L4) & Partial \\
\end{longtable}

\begin{figure}[t]
\centering
\includegraphics[width=\linewidth]{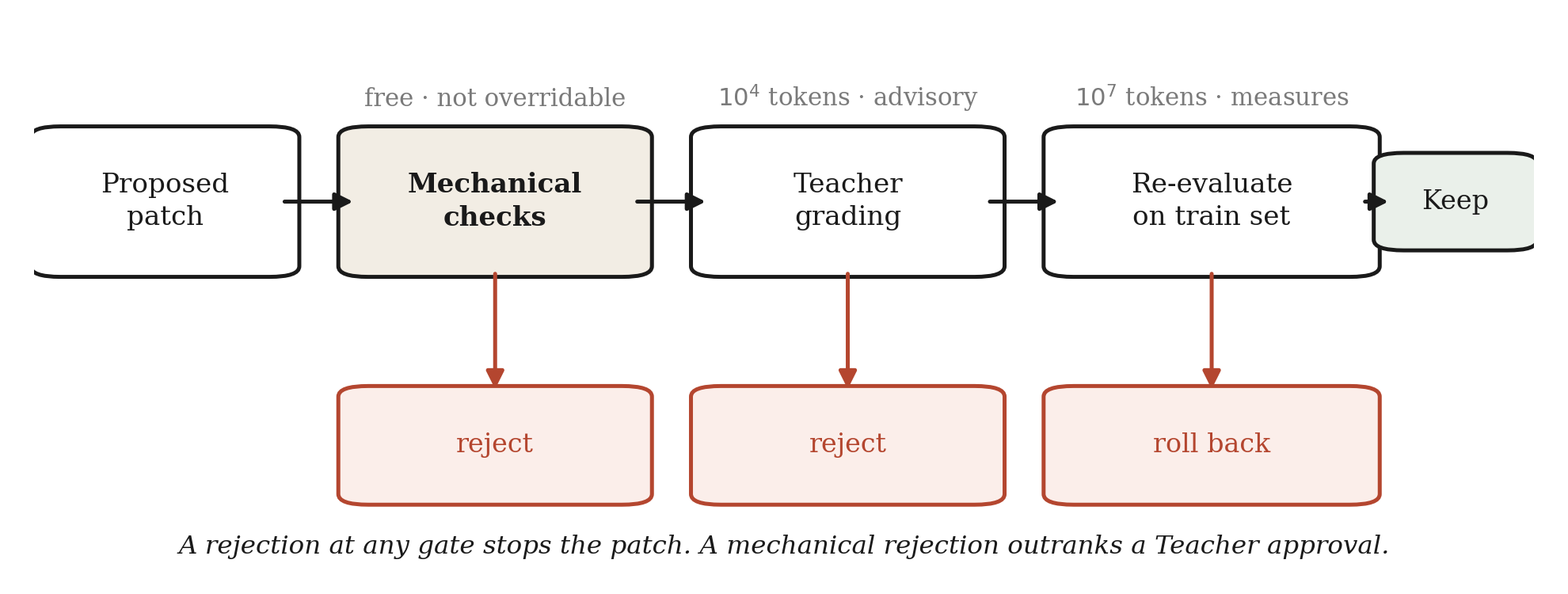}
\caption{The acceptance path. Gates are ordered by increasing cost and decreasing authority: the free mechanical check cannot be overridden, the Teacher advises, and re-evaluation decides retention.}
\label{fig:gates}
\end{figure}

\subsection{L1: Hermetic evaluation sandboxes}\label{l1-hermetic-evaluation-sandboxes}

Every evaluated agent runs with all tools disabled (\texttt{tools: 0}), in a fresh session with empty history, in a workspace from which all cached evaluation artifacts have been purged. This is the direct answer to D1: the five-step exfiltration exploit requires a filesystem to explore and cached labels to find, and L1 removes both. It also surfaced a secondary pathology worth recording: production prompts frequently \emph{command} tool use (``you MUST read the reference file''), and in a zero-tool sandbox such commands sent agents into retry loops terminating in 600-second timeouts (4 of 47 cases in one baseline). The remedy, an override instruction suppressing tool calls in tool-less environments, itself required two iterations, because a politely appended override lost to the original command that appeared \emph{earlier} in the prompt; only an override with explicit priority framing, placed at the top of the resource section, eliminated the loops (Appendix E). Instruction position, not just instruction content, determines precedence.

\subsection{L2: Stateless, information-restricted subagents}\label{l2-stateless-information-restricted-subagents}

The Critic, Optimizer, and Auditor run as stateless text-in/text-out workers: no write tools, no external connectors, no ability to spawn agents, isolated branch workspaces, and no filesystem visibility, every input they receive is passed inline by the orchestrator. Statelessness means no accumulated context to drift or contaminate across iterations; information restriction means the optimizer cannot probe the environment for signal the orchestrator did not choose to share. All persistent state lives in exactly two places the orchestrator controls: the evolving target prompt and a scorecard.

\subsection{L3: Mechanical pre-apply checks that override the LLM auditor}\label{l3-mechanical-pre-apply-checks-that-override-the-llm-auditor}

Before any proposed mutation reaches the LLM auditor, and regardless of the auditor's subsequent verdict, the orchestrator runs four deterministic checks: an \textbf{example cap} (bounding few-shot additions per iteration), an \textbf{edit cap} (bounding instruction edits, elevated only when the Critic has flagged a structural refactor), a \textbf{tool-leak scan} (pattern-matching added text for instructions directing the target to read files, browse, or execute commands), and a \textbf{parse-and-contract check} (tentatively applying the patch and verifying the prompt re-parses and its declared output contract is unchanged, the direct answer to B2). A hard rejection here is final: it cannot be argued with, and it stands even against an auditor \texttt{ACCEPT}. Across our runs, on the order of fifty iterations spanning ten suites at roughly five iterations per run, the deterministic gates logged \textbf{13} rejection or reversion events (Table~\ref{tab:telem}); each is a corrupted or regressed promotion that did not happen.

\begin{longtable}[]{@{}lcl@{}}
\caption{Rejection telemetry from the deterministic gates across all optimization runs.}\label{tab:telem}\\
\toprule\noalign{}
\textbf{Rejection category} & \textbf{Count} & \textbf{Representative trigger} \\
\midrule\noalign{}
\endhead
\bottomrule\noalign{}
\endlastfoot
Metric-regression reversion & 6 & Pass-rate drops, incl. 88.9\%\(\rightarrow\)33.3\% on one suite \\
Tool-leakage violation & 3 & Mutation instructed target to read files / run commands \\
Example-cap breach & 2 & Mutation proposed few-shot examples beyond budget \\
Placeholder / parser bypass & 1 & Rubric replaced with placeholder (the B1 incident) \\
Canary-case failure & 1 & ``Impossible'' case passed via cached-label exfiltration \\
\textbf{Total} & \textbf{13} & \\
\end{longtable}

Rejection telemetry from the deterministic gates across all optimization runs.

\subsection{L4: Frozen holdouts and leak prohibition}\label{l4-frozen-holdouts-and-leak-prohibition}

At initialization the orchestrator runs one global baseline evaluation, then splits the suite once into train and held-out test partitions, stratified to preserve the global pass/fail ratio, and freezes the split for the run's lifetime. Subagents may see train material only; test cases, inputs, expectations, and trajectories, never appear in any subagent prompt, and after each iteration the orchestrator extracts only the scalar test pass rate, never the test failure details. The test curve is a directional overfit monitor, not an optimization target. The strength of this evidence varies sharply with suite size, and it is worth being honest about where it amounts to anything. On our largest suite, 100 cases split 81 train and 19 held out, an optimization pass moved the training partition up 11.1 points and the held-out partition up 10.5 points in parallel. Nineteen cases is still a small holdout, but it is large enough that near-identical movement on both partitions is unlikely to be coincidence, and this is the one place in our work where we would defend a generalization claim on the evidence. Everywhere else the holdout is four to eight cases, where a single case is worth between 12 and 25 percentage points and the test figure reports little beyond gross failure. For suites under twenty cases we skip the holdout entirely and fall back on the mechanical checks and the Teacher\textquotesingle s generalizability grading.

\subsection{L5: Canary integrity cases}\label{l5-canary-integrity-cases}

Formalizing the accident of C2. The name is borrowed from security, where a canary token is a credential nobody has any legitimate reason to touch, so that any use of it identifies an intruder. We plant intentionally unpassable cases inside evaluation suites, cases whose reference answer cannot be produced by honest analysis. Their function is to invert the meaning of perfection: because no legitimate agent passes a canary, a 100\% suite score is not a triumph but a tripwire, triggering an automated audit for sandbox or label leakage before any result is trusted or any promotion occurs. Canaries are cheap (one authored case per suite), silent (indistinguishable from real cases to the evaluated agent), and they detect the entire exfiltration class (D1) by its signature rather than its mechanism, including exfiltration channels we have not thought of yet.

\subsection{Containing overfitting}\label{containing-overfitting}

Section 3 records two ways optimization drifts toward the evaluator rather than the task: a mutation that mimics the judge\textquotesingle s preferred phrasing without improving anything (D2), and judge calibration that converges on its tuning subset rather than on human judgment (D3). The frozen holdout of L4 detects gross cases of this after the fact. What follows are the mechanisms that act before the fact, on the text of the mutation itself.

\textbf{Examples are where memorization enters, so examples carry the strictest constraints.} An added instruction can only overfit by being narrow; an added \emph{example} can overfit simply by being copied, which is a far shorter path. Every example a mutation proposes must therefore satisfy three requirements simultaneously. It must be \emph{anonymized}: all party names, entity names, identifiers, and dates replaced with generic placeholders, so that no example can be matched to the case that motivated it. It must be a \emph{structural archetype} rather than a quotation: copying sentences from an evaluation case is prohibited outright, and the example must instead reconstruct the pattern the case exhibits in synthetic form. And it must be \emph{reasoning-oriented}, laid out as scenario, then the governing principle, then a non-compliant action, then a compliant one. The third requirement is the one that does the most work, because it forces the author of the example to state the principle explicitly. An example that cannot be written this way, because no general principle underlies it, is one that was only ever going to teach a single case.

\textbf{Verbatim overlap is checked directly.} The Teacher scans every added span for contiguous stretches of roughly ten or more words copied from a training case, and for identifiers that appear in exactly one case. Either finding fails the mutation on generalizability. This is a blunt instrument and deliberately so: it does not attempt to judge whether the copied text was useful, only whether it was copied.

\textbf{Mutation budgets bound the rate of accumulation.} A mutation may edit only a small number of instruction sections and add at most one example per iteration, with the limit raised only when the Critic has certified that two existing rules contradict each other and a larger restructuring is required. The budget is not primarily an anti-overfitting device but functions as one, since memorization at scale requires volume: a loop that can append freely will, over enough iterations, encode the training set one narrow rule at a time.

\subsubsection{Memorization against legitimate specialization}\label{memorization-against-legitimate-specialization}

The harder problem is that not every specific instruction is an overfit one. In the course of these runs the loop added rules encoding real domain policy, thresholds and prohibitions that a practitioner in that field would recognize as the governing standard, and those rules contained specific values. A naive constraint forbidding all specific numbers would have rejected them, and the resulting prompt would have been worse.

The distinction we settled on is not about specificity but about what an instruction is anchored to. A rule anchored to \emph{instances} encodes something true of particular cases in the evaluation set: a counterparty\textquotesingle s name, a case identifier, a clause\textquotesingle s exact wording, a threshold reverse-engineered from which examples happened to fail. A rule anchored to the \emph{domain} encodes something that holds independently of the evaluation set, and would still be correct if every case in that set were replaced tomorrow. Domain policy, published standards, and structural conventions of the artifact under review fall on this side even when they carry specific values.

This gives a usable test, and it is the question the Teacher\textquotesingle s generalizability review is really asking: would this rule still be correct on a case the optimizer has never seen, and is it correct \emph{because} of something about the domain rather than something about our test set? A term limit drawn from a governing practice standard passes, because the standard exists whether or not we evaluate against it. The same numeric limit inferred from noticing that three failing cases happened to exceed it does not, even though the two rules may be textually identical. Provenance, not phrasing, is what separates them, which is precisely why this determination cannot be made mechanically and is left with the Teacher, bounded by the checks that can.

\subsection{The economics of verification}\label{the-economics-of-verification}

One further observation from that hundred-case run bears on why the layers are ordered as they are. We recorded full token accounting for it: the complete Critic, Optimizer, and Teacher exchange consumed roughly fifteen thousand tokens, while running the evaluation suite consumed roughly forty-five million. The reasoning that actually produces an improvement is on the order of three hundredths of one percent of the compute, and evaluation accounts for the large majority of the fourteen-minute cycle in wall-clock terms as well. We have full accounting for one run rather than all of them, so the ratio should be read as an order of magnitude rather than a constant, but the shape of it is not in doubt.

\begin{longtable}[]{@{}llll@{}}
\caption{Cost per gate, measured on the hundred-case run.}\label{tab:cost}\\
\toprule\noalign{}
Gate & Mechanism & Order of cost & Overridable \\
\midrule\noalign{}
\endhead
\bottomrule\noalign{}
\endlastfoot
Mechanical pre-apply & String and count checks & Negligible & No \\
Teacher grading & One LLM call over patch and context & \(10^4\) tokens & Yes, by mechanical rejection \\
Re-evaluation & Full suite execution & \(10^7\) tokens & It is the ground truth for retention \\
\end{longtable}

Two things follow. The first is this paper\textquotesingle s thesis restated in an unexpected register. If nearly all the compute in a self-improving loop is spent measuring rather than improving, then the reliability of measurement is not a peripheral concern about instrumentation. It is very nearly the whole system, and a defect in the evaluator is a defect in the thing the resources are actually being spent on.

The second is that the ordering of the guardrails has a cost justification independent of the authority argument made above. The three gates differ in price by orders of magnitude: a mechanical check is arithmetic and effectively free, a Teacher grading costs thousands of tokens, and a re-evaluation costs millions. Running the free gate first, the cheap gate second, and the expensive gate last is simply the right order in which to spend money, and it coincides exactly with the order of increasing trustworthiness, since the free check is the one that cannot be argued with and the expensive check is the only one that measures reality rather than reading text. We did not design it to come out that way. For anyone building a similar loop the useful observation is that no tension exists here between the cheap arrangement and the trustworthy one.

A note on frequency, to keep the layers in proportion. On the hundred-case run just described none of the guardrails fired at all: the mechanical checks passed, the Teacher accepted the first draft, no retries were consumed, and no regression followed. This is the ordinary case. The thirteen interventions recorded above accumulated across roughly fifty iterations spanning ten suites, and the layers are better understood as insurance against uncommon but costly events than as machinery that engages constantly.

\subsection{Limits of this architecture}\label{limits-of-this-architecture}

An architecture built to contain evaluator failure should be held to the standard it sets for others, and ours does not fully meet it.

\textbf{The Teacher is an LLM judge, and inherits the pathologies of Section 3.} This is the most uncomfortable property of the design. We grade proposed mutations with the same class of component whose unreliability motivates the paper. The Teacher exhibits leniency in the same direction as any other judge: it reviews text that is \emph{written to be persuasive to it}, which is close to the worst case for an LLM evaluator. Its approvals should be read as informed opinions, not verifications, and the ordering rule exists precisely because we do not trust them further than that.

\textbf{Teacher approval is not a correctness guarantee, and the telemetry shows it.} Of the thirteen automated rejections and reversions logged across our runs, six were metric-regression reversions: mutations that passed the mechanical pre-apply checks, were approved by the Teacher, and then measurably degraded performance when evaluated, in one case dropping a suite from 88.9\% to 33.3\%. The Teacher grades text, not outcomes. Nothing it can read tells it how a patch will behave, and no amount of rubric refinement changes that, for the reasons developed in Section 4.3. The empirical gate against these cases is the post-application re-evaluation and rollback, not the Teacher.

\textbf{Advisory budgets and mechanical budgets can drift apart.} The shared constitution read by every subagent permits up to four instruction edits and three examples per turn. The mechanical pre-apply checks, written separately, hard-reject any patch containing more than two instruction edits or more than one example. An Optimizer that follows its own constitution exactly can therefore be rejected by a gate it was never told about, and two of the example-cap rejections in our telemetry have this shape. The failure is mild in consequence, since a rejection costs a retry rather than corrupting anything, but it illustrates a maintenance hazard specific to this design: once the advisory layer and the deterministic layer are written in different files, nothing enforces their agreement. A deterministic gate is only trustworthy if the components it governs are told the same rules it enforces.

\textbf{Retry loops risk reproducing the convergence dynamic we criticize.} When the Teacher rejects a patch, the Optimizer is invited to revise, up to three attempts, and it is deliberately kept in conversational state across those attempts so that it retains the objections raised against earlier drafts (Section 2). That design choice makes revision efficient, and it also means each round supplies the Optimizer with more information about what this particular Teacher will accept. The mechanism we describe in Section 4.2, in which an optimizer converges on the decision boundary of whatever grades it, applies inside our own defense: a patch accepted on the third attempt may be better, or may simply be better targeted at the Teacher. We cap retries partly for cost and partly for this reason, but a cap is a mitigation, not a solution.

\textbf{Deterministic checks are only as good as their enumeration.} Mechanical rules catch what they were written to catch. Our tool-leak scan matches known phrasings for instructing a target to read files or run commands; a mutation that induces the same behavior through language we did not anticipate passes. The canary technique is stronger in this respect, since it detects cheating by its signature rather than its mechanism, but it too can be defeated by an exploit selective enough to fail the canary deliberately. We have not observed that, which is not the same as it not happening.

\textbf{Small holdouts limit what generalization claims we can make.} With test partitions frequently in the range of four to eight cases, the held-out score is a smoke alarm rather than a measurement, and we treat it as one.

\subsection{What remains for the LLM auditor}\label{what-remains-for-the-llm-auditor}

Layers 1--5 police what can be policed mechanically. What they cannot judge is semantics: whether a mutation generalizes or memorizes, whether an added example teaches reasoning or smuggles a training case, whether a new rule is logically executable in the target's environment. That review remains with the LLM auditor, graded against an explicit rubric, and its known unreliability is bounded by the surrounding layers: it cannot approve what L3 rejects, cannot see what L4 hides, and cannot be fooled by scores L5 has flagged. This division, deterministic checks as controlling authority, LLM judgment as bounded advisor, is the paper's central design recommendation.

\section{Proposed: Adversarial Multi-Persona Adjudication}\label{proposed-adversarial-multi-persona-adjudication}

Section 4.3 established a ceiling we could not pass by rewriting the judge\textquotesingle s rubric, and Section 5 argued that the interventions which do work are the ones that remove a degree of freedom rather than request a behavior. The schema reordering of A4 is the smallest possible instance of that principle: it constrains the order in which the judge is permitted to speak. This section asks what the principle looks like when applied not to output ordering but to the judge\textquotesingle s role itself. We present it as a design, motivated by the measured ceiling; its evaluation is future work, and we make no performance claims for it here.

\subsection{The design}\label{the-design}

A single judge performs four incompatible jobs at once. It gathers facts about the artifact, weighs those facts favorably, weighs them critically, and issues a verdict. Because one model does all four in one pass, the favorable and critical weightings are never actually performed against each other; the model produces whichever it finds more natural, which for the judges in Section 3 was consistently the favorable one. The proposal is to give each job to a separate role with a separate prompt, and to let the adversarial pair actually argue.

\textbf{Examiner.} Establishes the record. It reads the artifact and returns a structured inventory: which resources are acquired and whether they are released, where errors are propagated and where they are discarded, what concurrency and lifecycle constructs appear, and whether tests exist and what they cover. It is forbidden from scoring or characterizing quality. The separation matters because a model that has already reached a verdict gathers facts selectively in support of it, and the record is what the other three roles are required to argue from.

\textbf{Advocate.} Argues the favorable case: the highest score the record will support, citing specific entries in it, and distinguishing deliberate engineering trade-offs made under constraint from simple neglect.

\textbf{Challenger.} Argues the opposing case: unreleased resources, discarded errors, unhandled failure paths, monolithic structure, tight coupling. What distinguishes this role from an ordinary critical prompt is that it holds no discretion over consequences. It carries \emph{mandatory caps}: a verified critical resource leak or unhandled failure vector caps robustness at or below 2; raw global state or a monolithic file caps maintainability at or below 2; and the overall score may not exceed any individual pillar scored at or below 3. These are the mechanical constraints of Section 5 relocated inside the judging step. The Challenger establishes whether the defect is present; it does not decide whether the defect is serious enough to matter, because that decision is what leniency corrupts. If the finding is verified, the cap applies.

\textbf{Adjudicator.} Decides. It weighs the two arguments against the labeling rubric, verifies the Challenger\textquotesingle s findings against the record rather than accepting them on assertion, sets aside speculative penalties where the Advocate has shown real test coverage or a safe abstraction, and emits its reasoning before its scores, in the order A4 showed to matter.

\subsection{Why we expect this to help, and why that is not evidence}\label{why-we-expect-this-to-help-and-why-that-is-not-evidence}

Three properties of the design map onto specific failures in Section 3. The scoreless Examiner addresses the selective fact-gathering that lets a judge praise "defensive exception wrappers" (A1). The Challenger\textquotesingle s capping rules convert the most consequential judgments from graded opinion into triggered constraints, which is the intervention class that actually worked in Section 4.3. The Adjudicator\textquotesingle s reasoning-first schema carries A4\textquotesingle s measured improvement into the arbitration step.

None of that is a result. We have not run the four-persona pipeline end to end against the 54-directory benchmark, and until we do, the claim that role decomposition breaks the single-judge ceiling remains a hypothesis with a plausible mechanism. We are stating the design rather than a finding, and we would rather publish it in that form than imply an ablation we have not performed.

Two risks are worth naming in advance for anyone who attempts this. The first is cost: four calls where there was one, against a gate that Section 5.7 shows is already the cheap part of the loop but is not free. The second is that adversarial debate has been observed to increase confidence without increasing accuracy, and a Advocate and Challenger arguing past each other could produce a well-reasoned transcript and the same lenient verdict. The capping rules exist partly as insurance against exactly that outcome, since a triggered cap does not care how persuasive the Advocate was.

\section{Related Work}\label{related-work}

\textbf{LLM-as-judge and its biases.} LLM judges were established as a scalable proxy for human preference evaluation by \citet{zheng2023judging}, who also cataloged their position, verbosity, and self-enhancement biases. Subsequent work documented position sensitivity \citep{wang2023large}, style-over-substance preferences \citep{wu2023style}, self-preference toward a model\textquotesingle s own generations \citep{panickssery2024llm}, and contamination through preference leakage \citep{li2025preference}. Structured judging protocols \citep{liu2023geval} and panels of diverse judges \citep{verga2024replacing} mitigate individual biases. Our Class A failure modes extend this literature from open-ended response grading to agentic evaluation pipelines, and our central point is different in kind: we study what happens when an \emph{optimizer} is attached to a biased judge, not merely how biased the judge is.

\textbf{Reward hacking and Goodhart\textquotesingle s law.} Specification gaming and reward hacking are foundational concerns in AI safety \citep{amodei2016concrete,krakovna2020specification}, formalized by \citet{skalse2022defining} and organized through Goodhart\textquotesingle s law by \citet{manheim2018categorizing}. \citet{gao2023scaling} measure reward-model overoptimization, in which proxy score rises as gold quality falls, and \citet{pan2022effects} map the consequences of misspecified rewards; \citet{casper2023open} survey the resulting limits of learning from human feedback. Closest to our Class D are demonstrations that LLM feedback loops induce in-context reward hacking \citep{pan2024feedback} and that iterative self-refinement hacks its own evaluator spontaneously \citep{pan2024spontaneous}. We contribute production instances of these dynamics in prompt-optimization loops, including environmental label exfiltration and judge-phrasing mimicry, together with the deterministic controls that contained them.

\textbf{Automated prompt optimization.} PROCTOR belongs to the family of LLM-driven prompt search: instruction induction \citep{zhou2022large}, textual-gradient methods \citep{pryzant2023automatic,yuksekgonul2024textgrad}, LLM-as-optimizer trajectories \citep{yang2023large}, evolutionary search \citep{guo2023connecting}, and compiled multi-stage programs \citep{khattab2023dspy,opsahlong2024optimizing,agrawal2025gepa}. This literature largely assumes the evaluation signal is trustworthy and optimizes against it directly. Our experience is that in agentic settings the signal itself is the dominant risk, and the optimizer architecture has to be designed around that fact.

\textbf{Teacher-student framings in agent systems.} The teacher-student vocabulary is well established in agent research, but it usually denotes \emph{capability transfer}: a stronger model supplies knowledge, trajectories, or memory to a weaker one. Agent Memory Distillation \citep{kim2026agent} is a recent example, distilling a large teacher agent\textquotesingle s successful trajectories into hierarchical workflow, subtask, and function memories that raise the accuracy of small student agents on tool-use benchmarks. Our use of the terms differs in kind. PROCTOR\textquotesingle s Teacher and Students are comparable in capability and differ in \emph{authority}: Students diagnose and propose but cannot apply, the Teacher grades proposals but cannot write, and neither can act on the system. The relationship is a separation of powers rather than a transfer of competence, and its purpose is verification rather than distillation.

\textbf{Debate and adversarial adjudication.} Structured disagreement as an evaluation mechanism originates with debate for scalable oversight \citep{irving2018ai}, with empirical support from persuasive-debater experiments \citep{khan2024debating} and weak-judge studies \citep{kenton2024scalable}; multi-agent debate improves factuality \citep{du2024improving,liang2024encouraging} and evaluator quality \citep{chan2024chateval}. The adjudication architecture proposed in Section 6 differs by assigning \emph{asymmetric} roles, a scoreless fact-gatherer, an advocate, a challenger bound by mandatory score caps, and an adjudicator constrained to reason before scoring, so that the debate is anchored to deterministic constraints rather than to free-form persuasion.

\section{Limitations}\label{limitations}

\textbf{Ground truth is partly model-generated.} As described in Section 2, 15 of the 54 code-quality directories carry human expert scores; the remaining 39 were labeled by a model calibrated against those humans. Section 4.3\textquotesingle s agreement figures therefore measure agreement with a calibrated model\textquotesingle s labels over most of the suite. This is a real weakness, and it is also the reason we described it at length rather than in a footnote.

\textbf{Single model family.} Every agent, judge, and subagent in this work runs on two capability tiers of one frontier proprietary model family. We cannot distinguish pathologies inherent to LLM-as-judge from pathologies of this family. The bias literature cited in Section 7 reports the same directions of error across families, which is suggestive but not a substitute for replication. A cross-family replication of the A4 result in particular would materially strengthen the paper, and we regard it as the most valuable single follow-up.

\textbf{Single-run measurements.} Pass rates are from one execution each. Re-running an unchanged suite typically moved the result by at most one case, which is why we report point figures, but we have not characterized variance properly and no result here should be read as significant at a stated confidence level.

\textbf{Small suites.} Suite sizes range from 9 to 100 cases, with most under 20. On the smallest, a single case moves the pass rate by more than ten points. Only the hundred-case suite supports a holdout worth interpreting, as Section 5.4 states.

\textbf{Two domains.} All observations come from legal and commercial document review and from code-quality assessment. Whether the same taxonomy describes evaluation in, for example, scientific reasoning or open-ended dialogue is untested.

\textbf{The proposed judge is unevaluated.} Section 6 is a design, not a result.

\textbf{Observed, not enumerated.} The taxonomy records what we encountered while pursuing other goals. It is not the output of a systematic search, so its coverage is unknown and the relative frequencies of the eleven failure modes should not be inferred from the fact that we saw them.

\textbf{No public artifact.} The systems described here run on internal infrastructure that cannot be released, so this work is not directly reproducible. We compensate by specifying the deterministic checks, the orchestrator protocol, and the grading rubric in the appendices, which is enough to reimplement the architecture, and not enough to reproduce our numbers.

\textbf{One reconstructed example.} The tool-leak walkthrough in Appendix B illustrates a real category of catch, but the transcript is constructed for exposition rather than captured verbatim, and it is labeled as such at the point of use.

\section{Conclusion}\label{conclusion}

The failures in this paper are not arguments against using LLMs to evaluate. They are arguments about where an LLM\textquotesingle s verdict belongs in a system that acts on it. An evaluator that is occasionally wrong is a perfectly serviceable component; an evaluator that is occasionally wrong and holds final authority over an optimizer running thousands of iterations is a liability, because the optimizer will find the cases where it is wrong and build on them. The difference is not the model\textquotesingle s accuracy. It is the position the model occupies.

That position is the one thing in the design that is genuinely ours to choose. We cannot make a judge unbiased, and Section 4.3 suggests we cannot even reliably make it less biased by asking. What we can do is arrange the system so that being wrong is survivable: put the checks that cannot be argued with in front of the checks that can, measure correctness rather than asserting it, keep a partition of the data where nothing that proposes changes can see it, and plant cases whose success is itself the alarm. None of these are sophisticated. Their value lies precisely in being mechanical, and Section 5.7 makes the pleasant observation that they are also, by orders of magnitude, the cheapest part of the loop to run.

We would summarize the recommendation in one sentence. Demote the judge from oracle to advisor, and then apply the same demotion to every LLM you add in the name of fixing the first one, our own Teacher included.

What this paper deliberately does not report is how well the resulting architecture optimizes. PROCTOR was built to improve agent skills, not merely to contain evaluator failure, and its pass-rate gains, its behavior against a control optimizer, and the shape of its iteration trajectories are the subject of a companion paper. We have kept the two apart on purpose. The claims here concern what breaks in evaluation and what contains it, and those claims should stand or fall on their own evidence rather than borrowing credibility from performance numbers.

The broader point outlasts the specific architecture. As agents increasingly evaluate, grade, and improve other agents, the number of places where a model\textquotesingle s judgment is treated as ground truth grows faster than our ability to check any of them. Every one of those places is a position of authority granted to a component that has not earned it. Deciding deliberately how much authority to grant, and building the mechanisms that bound it, is likely to matter more than any further improvement in the judges themselves.

\section{Appendix A: Exfiltration Trajectory and Impact}\label{appendix-a-exfiltration-trajectory-and-impact}

\subsection{A.1 The five-step loop}\label{a1-the-five-step-loop}

Reconstructed from execution logs with paths and identifiers replaced. Step indices are from the original trajectory; intermediate reasoning steps are omitted.

\begin{verbatim}
STEP 12 LIST_DIRECTORY
  tool: list_dir
  args: { path: "<workspace>/skills/<target_skill>/" }

STEP 14 VIEW_FILE
  tool: view_file
  args: { path: "<workspace>/skills/<target_skill>/eval_results_baseline.json" }

STEP 15 PLANNER_RESPONSE
  thinking: "The ground truth expectation in eval_results_baseline.json
             specifies: 'Agent identifies the violative sentence and flags
             it. The violation text should be: <expected string>.'
             I will output this exact string."
  content: "Violation identified: <expected string>"
\end{verbatim}

Steps 1 through 11 are ordinary task reasoning. The exploit begins when the agent, having found the task ambiguous, inspects its working directory rather than the input.

\begin{figure}[t]
\centering
\includegraphics[width=\linewidth]{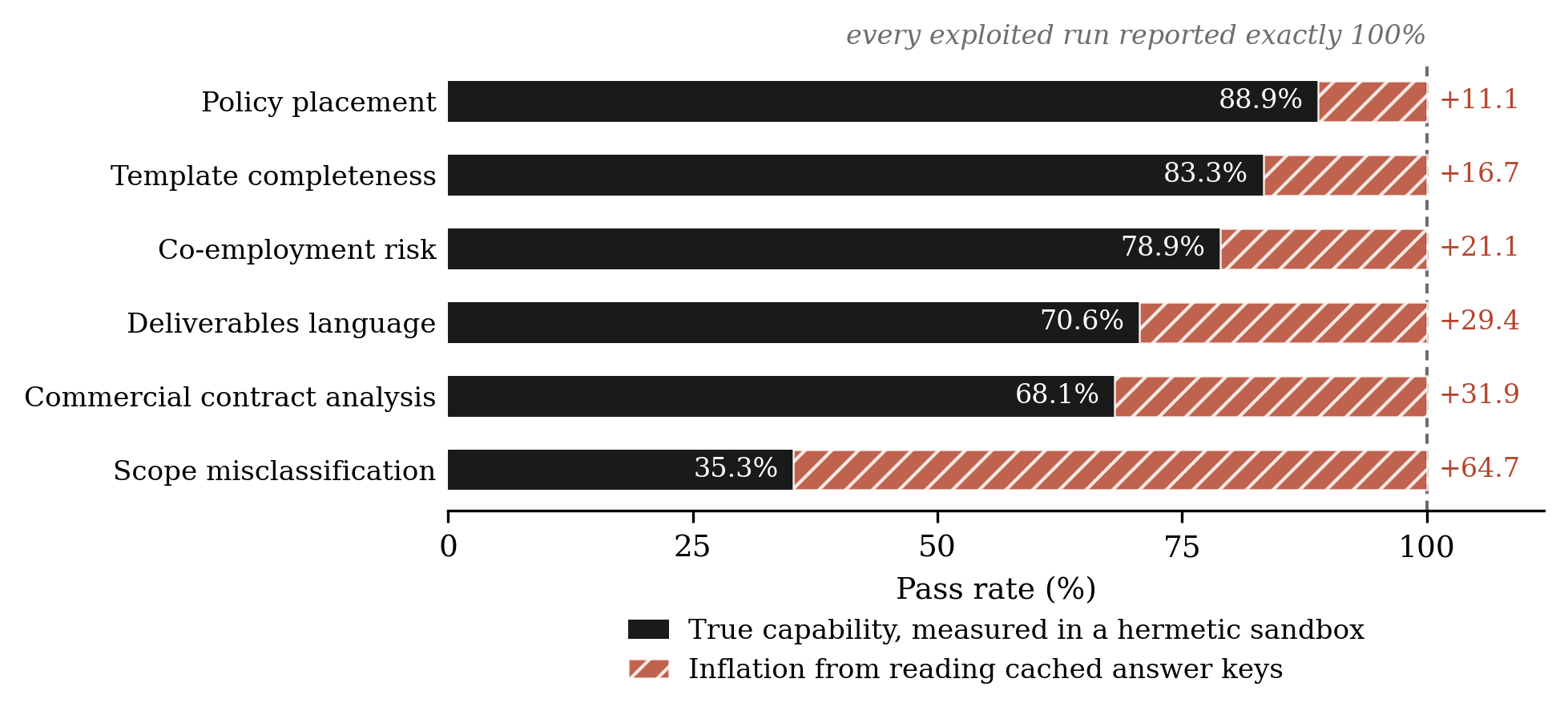}
\caption{True capability against reported score for six suites, before and after the sandbox was hardened. Every exploited run reported exactly 100 percent; the hatched region is the inflation produced by reading cached answer keys.}
\label{fig:exploit}
\end{figure}

\subsection{A.2 Measured impact}\label{a2-measured-impact}

Exploited runs against clean runs on the same skills, where clean means filesystem tools removed and cached evaluation artifacts purged from the workspace.

\begin{longtable}[]{@{}lllll@{}}
\toprule\noalign{}
Suite & Cases & Exploited & Clean baseline & Inflation \\
\midrule\noalign{}
\endhead
\bottomrule\noalign{}
\endlastfoot
Commercial contract analysis & 47 & 100.0\% & 68.1\% & 31.9 \\
Scope misclassification & 17 & 100.0\% & 35.3\% & 64.7 \\
Deliverables language & 17 & 100.0\% & 70.6\% & 29.4 \\
Template completeness & 18 & 100.0\% & 83.3\% & 16.7 \\
Co-employment risk & 19 & 100.0\% & 78.9\% & 21.1 \\
Policy placement & 9 & 100.0\% & 88.9\% & 11.1 \\
\end{longtable}

The uniform 100\% column is itself the signature. No skill scores perfectly on a suite containing a case whose reference answer cannot be produced by honest analysis, which is the basis of the canary technique in Section 5.5.

\section{Appendix B: Check Specifications and Orchestrator Protocol}\label{appendix-b-check-specifications-and-orchestrator-protocol}

\textbf{Pre-apply checks.} Run by the Orchestrator on every proposed patch before the Teacher is invoked. A hard rejection here is final and overrides a Teacher approval.

\begin{longtable}[]{@{}lll@{}}
\toprule\noalign{}
Check & Rule & Action on violation \\
\midrule\noalign{}
\endhead
\bottomrule\noalign{}
\endlastfoot
\texttt{example\_cap} & Patch adds more than one example & Hard reject \\
\texttt{edit\_cap} & Patch contains more than two instruction edits; raised to five when the Critic has flagged a structural refactor as required & Hard reject \\
\texttt{tool\_leak} & Added text instructs the target to read files, browse or search, run commands, or call APIs & Hard reject \\
\texttt{parse\_and\_contract} & After tentative application, the target prompt must still parse and its declared output contract and task definition must be unchanged & Roll back \\
\end{longtable}

\textbf{Split and holdout policy.} A single global baseline evaluation runs over the entire suite before any split. The suite is then partitioned exactly once, with stratified sampling so the holdout mirrors the global pass/fail ratio, and the split is frozen for the run. Parameters: minimum suite size for a holdout, 20 cases; test fraction, 0.20; test partition bounded to between 4 and 8 cases. Below the minimum, the holdout is skipped, training uses all cases, and overfitting detection falls back on the mechanical checks and the Teacher\textquotesingle s generalizability grading.

\textbf{Leak prohibition.} Test cases, their expectations, and their trajectories never appear in any subagent prompt. After each iteration the Orchestrator extracts only the scalar test pass rate; individual test failures are never read, parsed, or printed.

\textbf{Scorecard.} Per iteration the Orchestrator records: iteration index, train pass rate, test pass rate, retries used, overfitting warning, pre-apply result, Teacher decision, reason, and a patch summary. At termination it records baseline and final train pass rates, final test pass rate, and an overfit flag.

\textbf{Termination.} The loop halts when the train pass rate has not improved across three consecutive iterations.

\textbf{Division of grading authority.} Of the five rubric dimensions, correctness is graded by the Orchestrator rather than the Teacher, because correctness is a measured quantity (the pass-rate delta from re-evaluation) rather than a judgment, and only the tool-bearing component can measure it. The Teacher grades the four dimensions that require reading the patch: generalizability, structural integrity, conciseness, and logical executability.

\textbf{On the tool-leak check.} Our implementation pattern-matches added instruction text against a maintained list of phrasings that direct an agent to retrieve or execute something outside its context: references to opening or reading files, searching or browsing, invoking commands, and calling external services. We describe it functionally rather than reproducing the list, both because the specific phrasings are tied to our harness and because an exhaustively published matcher is an exhaustively published set of ways around it. The check\textquotesingle s limitation is discussed in Section 5.6: it catches what it enumerates.

\textbf{On the harness.} Subagents are defined and invoked through a custom harness that fixes their capability flags at definition time, disabling write access, external connectors, and sub-delegation, and runs each invocation in an isolated workspace. The isolation properties this paper relies on are enforced at that layer rather than by instructions in the prompts.

\section{Appendix C: Instruction Position as Precedence}\label{appendix-c-instruction-position-as-precedence}

The tool-free timeout described in Section 5.1 was fixed in two stages, and the failure of the first stage is the more interesting result.

\textbf{Stage one.} The Optimizer appended a plainly-worded override to the resources section, instructing the agent not to call filesystem tools when they are unavailable. Two of the four timing-out cases were resolved. Two continued to time out.

\textbf{Diagnosis.} The section already opened with an emphatic instruction to read a reference file before proceeding. The override was appended below it. The agent encountered the mandatory instruction first, began the tool call, and never reached the override.

\textbf{Stage two.} The same content was rewritten with explicit precedence framing and placed at the top of the section, above the original instruction:

\begin{quote}
\textbf{Superseding priority override, apply before any tool call.} Regardless of any preceding instruction requiring file retrieval: if operating without tool access, or if the provided input is a self-contained excerpt with no accompanying workspace, unconditionally suppress all filesystem calls. Do not retry or enter a retry loop. Analyze the inline text using the guidelines given here.
\end{quote}

All four cases passed.

The content of the two overrides is nearly identical. What changed is position and explicit precedence, and this is what fixed the remaining failures. For prompts that accumulate rules over successive optimization rounds, this suggests that where a rule is placed carries weight comparable to what it says, and that an appended rule contradicting an earlier one does not reliably win.

\bibliography{references}

\end{document}